\documentclass{article}

\usepackage{epsf}
\usepackage{complex-systems}
\usepackage{graphics}

\begin{document}

\title{Segmentation and Context of Literary and Musical Sequences}

\author{\authname{Dami\'an H. Zanette}
\\[2pt]
\authadd{Consejo Nacional de Investigaciones Cient\'{\i}ficas y T\'ecnicas}\\
\authadd{Centro At\'omico Bariloche and
Instituto Balseiro}\\
\authadd{8400 Bariloche, R\'{\i}o Negro, Argentina}}

 \maketitle

\markboth{Dami\'an H. Zanette}{Segmentation of Literary and Musical
Sequences}

\begin{abstract}
We test a segmentation algorithm, based on the calculation of the
Jensen-Shannon divergence between probability distributions, to two
symbolic sequences of literary and musical origin. The first
sequence represents the successive appearance of characters in a
theatrical play, and the second represents the succession of tones
from the twelve-tone scale in a keyboard sonata. The algorithm
divides the sequences into segments of maximal compositional
divergence between them. For the play, these segments are related to
changes in the frequency of appearance of different characters and
in the geographical setting of the action. For the sonata, the
segments correspond to tonal domains and reveal in detail the
characteristic tonal progression of such kind of musical
composition.
\end{abstract}

\section{Introduction}
\label{intro}

Natural systems store, retrieve, transmit, and exchange information
by means of a wide variety of mechanisms. This diversity is apparent
if we compare, for instance, the genetic code, a bird song, and a
written text. Genetic information, which contains the instructions
to build up proteins out of the chemical substrate inside living
cells, is processed at the level of DNA molecules, involving
chemical reactions and elementary quantum-mechanical interactions.
At a much more macroscopic level, the elaboration, emission, and
reception of a bird song --a courtship call, for instance--
activates a series of intermingled processes which involve at least
several regions of the brain, the vocal tract, and the ear. A
written text, characteristic of some human languages, presupposes a
conventional agreement for the symbolic representation of meaningful
phoneme sets --the words. Information processing through written
language always requires, to some extent, resorting to artificial
human-made tools, from pen and paper to computers.

It is remarkable that, in spite of the essential differences in
nature between the basic phenomena which sustain information
processing in its various forms, some universal features can still
be found to underlie the way in which information is organized in
different systems. One of these features, directly related to the
sequential character of information processing, is associated with
the possibility of translating pieces of information into ordered
series of symbols. Identifying the proper information units to be
translated into individual symbols may be easier --although rarely
trivial-- in some systems such as DNA molecules or human languages,
and more complicated in others, such as animal calls. A complex
message conveying information simultaneously at many sensorial
layers, such as a musical piece, may even admit several very
different symbolic codings, emphasizing different informational
aspects of the same message. However, once a translation code has
been agreed upon, symbol streams can be generated and analyzed as a
consistent representation of the information contents of the system
under study.

Due to the very essence of information, a symbolic sequence
representing a message is expected to possess nontrivial structural
properties at many length scales. Over short ranges, such properties
are governed by a set of rules which govern the combination of
adjacent symbols or, in other words, by grammar. At this level, the
structure of many human languages is reasonably well understood,
which has also led to the development of an abstract mathematical
theory of (natural and artificial) languages. In the case of DNA
molecules, localized patterns are directly related to the coding of
individual proteins, which control in turn specific phenotypic
traits in the organism. For other informational systems, on the
other hand, even short-range structures have unidentified functional
roles. Over larger scales, long streams of human language should
reveal patterns associated with meaning, context, and style
\cite{entro}. These global properties are also expected to appear in
long musical compositions \cite{mus}. Long-range structures in DNA,
extending over several thousand nucleotides, are by far less
understood and their interpretation is controversial
\cite{Amato,Karlin}.

While extracting grammar rules from the symbolic sequences of any
kind of language requires a detailed inspection of the local
configuration of symbols, regularities at larger scales are better
detected by statistical means. DNA  \cite{Kaneko,Peng,Voss} and
language sequences \cite{l1,l2,Pury} have already been analyzed in
this way, revealing long-range correlations which, due to their
power-law dependence on length, seem to point out the presence of
some kind of overall fractal structure. In the case of DNA, such
correlations have been ascribed to the presence and organization of
``patches'' with different nucleotide composition \cite{Kaneko94}.
To disclose and characterize this patchiness, a segmentation
algorithm for DNA sequences has been proposed on the basis of
maximizing the compositional difference between the resulting
segments \cite{hisp}, as measured by the Jensen-Shannon divergence
\cite{JS,hispp}. The method is aimed at detecting domains within
which the relative frequencies of the four nucleotides may sensibly
differ from other portions of the sequence. Besides giving an
acceptable definition of ``patch'', by its identification with a
segment, the recursive application of the segmentation algorithm has
made it possible to reveal power-law distributions for the segment
length \cite{hisp}. This broad distribution of patch sizes has
recently been proposed as a fingerprint of complexity in the
analyzed symbolic sequence \cite{compl}.

The aim of this paper is to illustrate the result of applying the
segmentation algorithm to other classes of symbolic sequences,
specifically, to sequences of linguistic or musical origin. The
possibility of detecting segments with different composition in this
class of sequences is particularly appealing, since such segments
may  be identified with contextual domains. Understanding context as
the emergent property of a message which sustains its
intelligibility in the long run \cite{mus}, those domains would
correspond to portions where the distinct internal distribution of
perceptual elements determines specific contextual features. Which
of these features are detected by segmentation depends, as shown
below, on the choice of the collection of symbols out of which the
sequence under study is built up. In the examples presented here, we
focus on the variation in the distribution of characters along a
theatrical play, and in the domains of tonal context in a musical
composition. In the next section, the segmentation algorithm is
introduced. Sections \ref{lit} and \ref{mus} are respectively
devoted to the segmentation analysis of sequences of literary and
musical origin. Results, as well as limitations and perspectives of
the method, are discussed in the last section.

\section{Compositional segmentation of symbolic sequences}
\label{anal}

As advanced in the Introduction, the segmentation algorithm used
below is based on the calculation of the Jensen-Shannon (JS)
divergence \cite{JS}. Given two probability distributions $\bf p$
and $\bf q$  over $k$ discrete states, ${\bf p}= (p_1,p_2,\dots,
p_k)$ and ${\bf q}= (q_1,q_2,\dots, q_k)$, the JS divergence defines
a distance between the two distributions, as
\begin{equation}
D[{\bf p},{\bf q}] = H[w_p {\bf p}+w_q {\bf q}]-w_p H[{\bf p}] -w_q
H[{\bf q}],
\end{equation}
where
\begin{equation}
H[{\bf p}]= -\sum_{i=1}^k p_i \log_2 p_i
\end{equation}
is Shannon's entropy for the distribution ${\bf p}$. The positive
weights $w_p$ and $w_q$ satisfy $w_p+w_q=1$, and the linear
combination $w_p {\bf p}+w_q {\bf q}$ is a probability distribution
with components $(w_p p_1+w_q q_1,w_p p_2+w_q q_2, \dots ,w_p
p_k+w_q q_k)$. It is assumed that the distributions ${\bf p}$ and
${\bf q}$ are normalized to unity, so that $\sum_i p_i=\sum_i
q_i=1$. This implies that $w_p {\bf p}+w_q {\bf q}$  is also
normalized.

As a distance between two distributions, the JS divergence verifies
some desirable properties. In particular, it is symmetric: $D[{\bf
p},{\bf q}] =D[{\bf q},{\bf p}]$. Moreover, $D[{\bf p},{\bf q}] >0$
for all ${\bf p}\neq {\bf q}$, and $D[{\bf p},{\bf q}] =0$ if and
only if ${\bf p}= {\bf q}$.

Suppose now to have a sequence of length $N$, whose elements are
taken from an ``alphabet'' of $k$ different symbols $\{
s_1,s_2,\dots,s_k\}$. Consider the two subsequences given by the
first $n$ elements of the sequence and the remaining $N-n$ elements.
In the first subsequence, of length $n$, the frequency of the symbol
$s_i$ is $f_i=n_i/n$, where $n_i$ is the number of occurrences of
$s_i$ in that subsequence. Note that, since $\sum_i n_i=n$, the
distribution ${\bf f}=(f_1,f_2,\dots, f_k)$ is normalized to unity.
Proceeding in the same way for the second subsequence, the frequency
of symbol $s_i$ is $g_i=(N_i-n_i)/(N-n)$, where $N_i$ is the total
number of occurrences of $s_i$ in the whole sequence. Again, the
distribution ${\bf g}=(g_1,g_2,\dots, g_k)$ is normalized to unity.

The distributions $\bf f$ and $\bf g$ characterize the frequencies
of symbols in each subsequence, and their JS divergence  gives a
measure of how different they are. In other words, $D[{\bf f},{\bf
g}]$ quantifies the compositional difference between the two
subsequences. The weight of each distribution in the JS divergence
is usually taken to be proportional to the length of the
corresponding subsequence \cite{hisp}, i.e. $w_f=n/N$ and
$w_g=(N-n)/N$. We thus calculate
\begin{equation}
D_n[{\bf f},{\bf g}] = H\left[ \frac{n}{N} {\bf f}+ \left(
1-\frac{n}{N} \right){\bf g}\right]-\frac{n}{N} H[{\bf f}] - \left(
1-\frac{n}{N} \right) H[{\bf g}].
\end{equation}
The subindex $n$ emphasizes the fact that, generally, the value of
the JS divergence depends on where the whole sequence is divided
into the two subsequences. Note that, in the trivial limits $n=0$
and $n=N$, we have $D_0=D_N=0$. Varying $n$ from $1$ to $N-1$, the
point $n_{\max}$ where $D_n$ reaches a maximum can be readily
detected. At that point, the compositional divergence of the two
subsequences is largest: $D_{\max} \equiv D_{n_{\max}} \ge D_n$ for
all $n$. The first segmentation is therefore performed at
$n_{\max}$, and two segments with maximal compositional divergence
result.

Once the first segmentation step has been achieved, the algorithm
can be applied to the resulting segments. Further iteration the
process will produce, at successive segmentation levels, four,
eight, sixteen, $\dots$ segments of decreasing length. At any level,
each pair of contiguous segments will represent the optimal division
of one the segments of the previous level with respect to their
compositional divergence. Therefore, the segmentation obtained at
each level contains increasingly detailed information about
variations in the frequency of different symbols along the original
sequence.

In principle, since $D_n$ will always reach a maximum within any
segment of length larger than or equal to $3$, the segmentation
process can be iterated until segments are reduced to a minimal
length of $2$. It is obvious, however, that well before the original
sequence becomes atomized into such minimal segments, the resulting
segmentation will cease to bear any information related to
long-range compositional variations. When segments are short enough,
the algorithm will be rather detecting the effect of random-like
local fluctuations in the distribution of symbols, with no
significant connection with  long-range structures.

A criterion for halting the iteration of the segmentation algorithm
can be established by comparing the maximal value of the JS
divergence along the (sub)sequence being processed with the typical
value expected for $D_n$ in a random sequence of identical length
and the same symbol frequencies \cite{hisp}. For instance, we can
assume that the maximum in the JS divergence is significant if it is
larger than the sum of the average value of $D_n$ plus its mean
dispersion in the random sequence. Unfortunately, evaluating
statistical properties of the JS divergence, such as its mean value
and dispersion, is not an easy task even for a random uncorrelated
sequence --although approximations have been provided \cite{hispp}.
In our analysis of linguistic and musical sequences, on the other
hand, we have adopted a heuristic approach which, operationally,
turns out to be quite convenient. After calculating the JS
divergence along the (sub)sequence to be segmented and detecting its
maximum $D_{\max}$, we randomly shuffle the (sub)sequence and
calculate $D_n$ again. The typical result, as a function of $n$, is
a kind of noisy signal whose average value and mean dispersion can
be calculated in the standard way:
\begin{equation}
\bar D = \frac{1}{N_s} \sum_{n=1}^{N_s} D_n, \ \ \ \ \ \ \sigma^2_D
= \frac{1}{N_s} \sum_{n=1}^{N_s} (D_n-\bar D)^2.
\end{equation}
Here, $N_s$ is the length of the (sub)sequence. The above-stated
criterion would imply that the segmentation is significant  if
$D_{\max}> \bar D+\sigma_D$.

In the following sections, we illustrate the application of these
methods to two sequences, respectively, of literary and musical
origin. Of course, a preliminary, not necessarily trivial task is to
decide how to extract a symbolic sequence out of a written text or a
musical composition. Once this point is defined, however, the above
algorithm can be straightforwardly applied to segment the sequence,
the significance of segmentation can be evaluated, and results can
be discussed in the perspective of the contents of the literary and
musical works under consideration.

\section{Segmentation of a sequence of literary origin}
\label{lit}

A written text can naturally be thought of as a sequence of symbols.
At the level of the semantic contents of the message, i.e. of its
meaning, individual symbols can be identified with single words.
This choice immediately posses the problem that, in any text of
substantial length, the number of different symbols is enormous. For
instance, a literary work of the size of Dickens's David Copperfield
or of Cervantes's Don Quijote, whose length spans a few hundred
thousands words, uses some 20,000 to 30,000 different words. On the
average, thus, each word appears some ten times all over the work.
The evaluation of probabilities from such small frequencies would
hardly be statistically significant. There is moreover the problem
that those words whose frequency is high enough as to allow for a
satisfactory determination of a probability distribution, are
usually irrelevant in defining the text's semantic contents. The
most frequent words in any human language --in English, for
instance, the words {\sl the}, {\sl and}, {\sl a}, {\sl of}, etc.--
convey no specific information on the contextual domain. We conclude
that a sequence of symbols derived from a written text must be built
up in a different manner.

In a previous work \cite{entro}, it has been shown that, over a
large literary corpus, different words are distributed more or less
homogeneously depending on their grammatical function. Nouns --and,
most particularly, proper nouns-- are much more specific to
localized domains of the corpus than, for instance, verbs or
adverbs. In other words, the former are distributed less
homogeneously that the latter, thus conveying more information about
the local context. In the frame of the present study, this property
can be exploited by choosing as a set of symbols a selection of
those words with larger specificity, and building up a sequence with
just such symbols. In the following, we apply these ideas to extract
a sequence of symbols from Shakespeare's play Othello \cite{oth}.

From the viewpoint of the text structure, a play is a succession of
(usually not very long) speeches, alternatively uttered by the
various characters. Let us call each one of these successive
speeches a {\it line}. Let us moreover associate a symbol to each
character. If each line is now assigned the symbol of the character
who utters it, the succession of lines along the play defines a
sequence of symbols which represents the progressive participation
of the different characters. For instance, Othello's first act,
which opens with a dialog between Roderigo (R) and Iago (I), with
the later entrance of Brabantio (B), is represented by the sequence
RIRIRIRIRIRIRIBRIBI... The choice of a sequence built up by these
symbols would lead the segmentation process to reveal the presence
of domains with relatively different relevance of the various
characters. It thus focuses on the specific component of context
determined by character presences and absences as the action
proceeds.

Shakespeare's Othello has $23$ ``actors'' (characters) \cite{oth},
and spans $1173$ lines. Therefore, each symbol in the sequence
appears an average of some $50$ times. However, the five most
important characters utter a total of more than $900$ lines. These
are Othello ($274$ lines), Iago ($272$ lines), Desdemona ($166$
lines), Cassio ($110$ lines), and Emilia ($103$ lines); the next
most important character reaches less than $60$ lines. Along the
play, frequency changes in the appearance of the five main
characters on scene dominate the compositional variation between
different parts of the sequence and, thus, have an essential role in
determining the result of segmentation. The role of other characters
is secondary --though, as we show below, they have an effect at
certain segmentation levels.

The play is divided into five acts. The first act takes place in
Venice, while the remaining of the action is in Cyprus. Since not
all the characters migrate from Venice to Cyprus, we expect that the
change of geographical environment be associated with a variation in
the relative distribution of characters, which may be detected by
segmentation. More subtle location transitions --such as, for
instance, between different places in Venice during the first act--
may also induce compositional variations in the sequence. Each act
is divided into two to four scenes of rather disparate lengths.
Unlike the classical convention which prescribes a change of scene
each time a character enters or exits, Shakespeare's division of
acts into scenes is freer. Still, scene beginnings and ends are
usually associated with major changes in the action and its agents.
Segmentation could therefore be also related to the division in
scenes.

\begin{figure}
\resizebox{\columnwidth}{!}{\includegraphics*{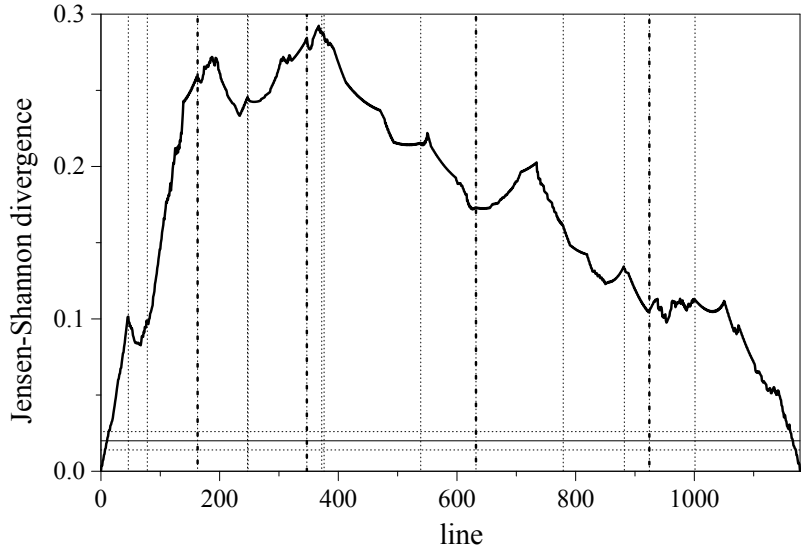}}
\caption{Jensen-Shannon (JS) divergence for the symbolic sequence
obtained from Shakespeare's Othello, as a function of the line
number. Vertical dotted lines stand for the division into acts
(bold) and scenes (light). The horizontal full line  corresponds to
the average value $\bar D$ of the JS divergence of a random sequence
with the same length and symbol frequencies. Dotted lines correspond
to the values $\bar D \pm \sigma_D$, where $\sigma_D$ is the mean
dispersion of the JS divergence over the random sequence.}
\label{oth1}
\end{figure}

Figure \ref{oth1} shows the JS divergence for the whole sequence of
$1173$ lines. Vertical full and dashed lines indicate the division
into acts and scenes, respectively. Rather unexpectedly, the maximum
in the JS divergence, at line $367$, does not coincide with any of
those divisions. It is therefore not revealing a mere localized
exchange of characters or an episodic turning point in the action,
but rather a more global --and perhaps more subtle-- change in the
characters' relevance on stage. In fact, an analysis of the
progressive appearances of individual characters shows that, in the
first third of the play, the protagonists Othello and Desdemona
utter only one sixth of their total number of lines. During this
first third, both of them appear episodically in most occasions,
usually in the company of several other characters, and do not
participate of long dialogues. In the remaining two thirds of the
play, on the other hand, Othello and Desdemona are practically
omnipresent --either together or not-- and, except for some short
scenes, clearly dominate the action progression. The maximum in the
JS divergence corresponds to this major, but not necessarily
obvious, change in the participation of the protagonist couple.
\begin{figure}
\resizebox{\columnwidth}{!}{\includegraphics*{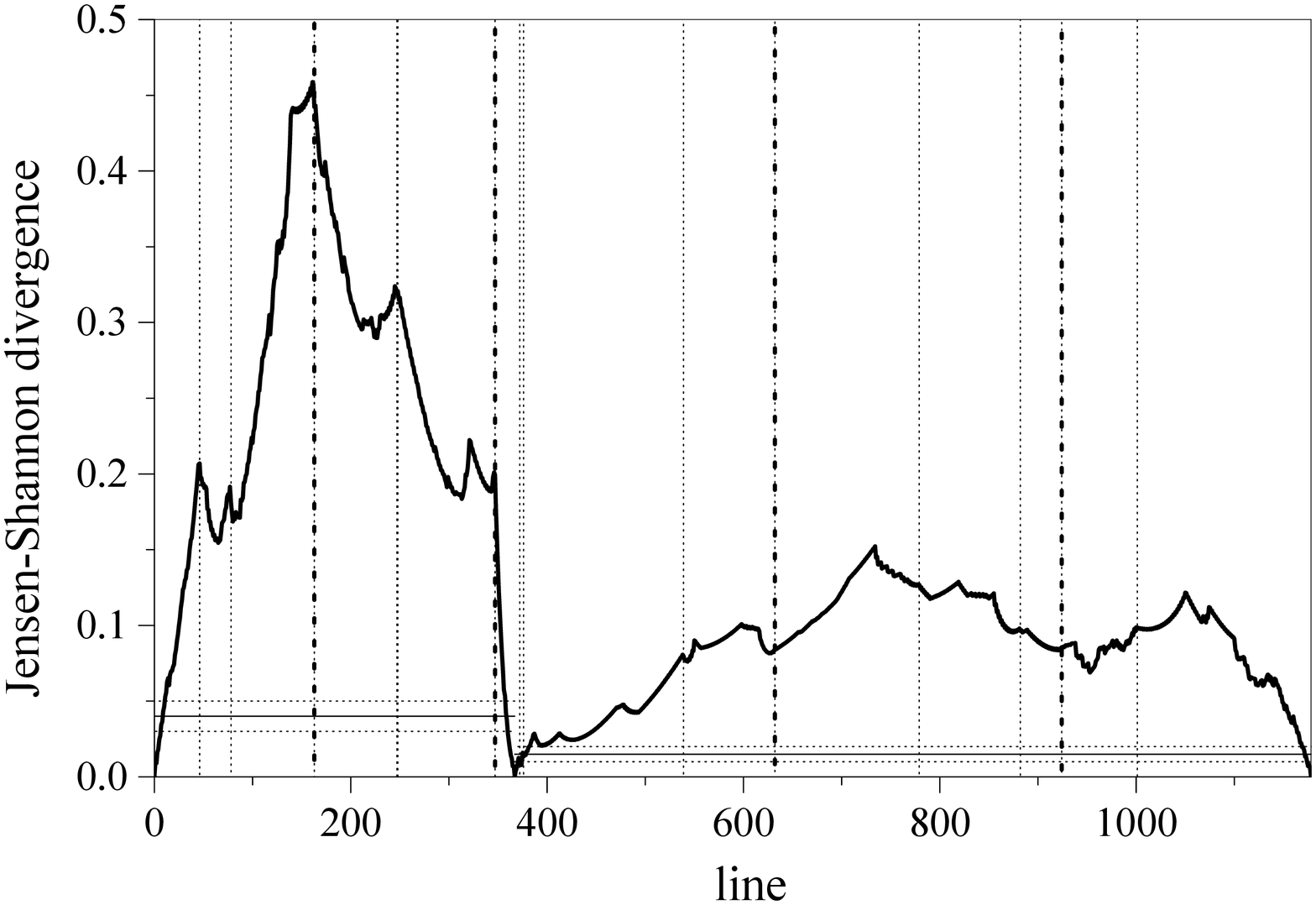}} \caption{As
in Figure \ref{oth1}, for the two segments resulting after the first
segmentation.} \label{oth2}
\end{figure}

The JS divergence after the first segmentation step is shown in
Figure \ref{oth2}. In the first part, it shows a sharp peak at line
$161$, at the precise end of Act I. As mentioned above, in the
transition form the first to the second act the action is
transferred from Venice to Cyprus. Correspondingly, several
relatively dominant characters of Act I --such as Brabantio, the
Duke and the Senators-- definitively disappear from the stage. Thus,
through this sudden change in the appearance frequency of otherwise
minor characters, the JS divergence detects the shift in the
geographical setting. In the second segment, the JS divergence
reaches its maximum at line $734$. This line does not coincide with
a scene transition, but lies in a zone of considerable frequency
changes in two characters, Cassio and Desdemona. After an appearance
at the beginning of Act IV, Cassio is absent for the rest of the
act. Desdemona, on the other hand, dominates vast sections from its
second scene.

\begin{figure}
\resizebox{\columnwidth}{!}{\includegraphics*{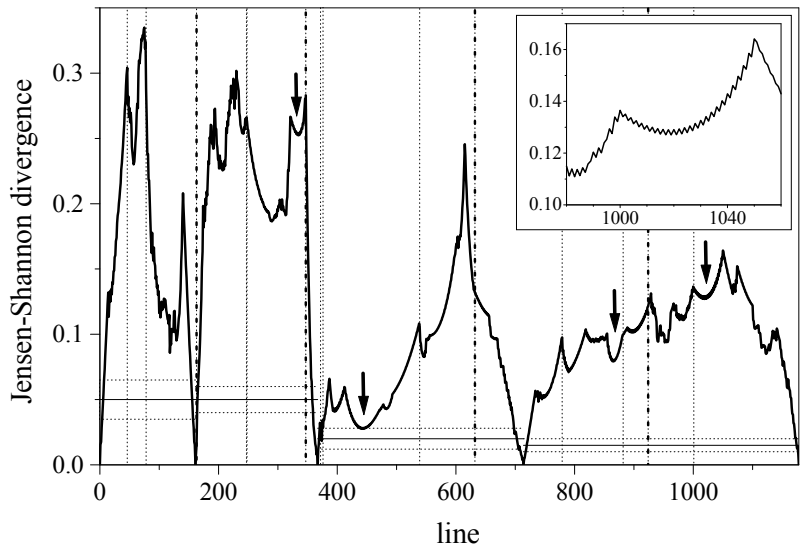}} \caption{As
in Figure \ref{oth1}, for the four segments resulting after the
second segmentation. Arrows indicate some of the U-shaped structures
corresponding to two-character dialogues. One of the is shown in
detail in the insert.} \label{oth3}
\end{figure}

The next segmentation level is depicted in Figure \ref{oth3}. In the
fist segment, the maximum occurs at line $75$, just at the end of
the second scene of Act I. Again, this maximum is detecting a
setting transition. While the two first scenes take place in a
Venetian street, the third scene unfolds in a ``council chamber''
--presumably, in the Ducal Palace-- with the ensuing change in the
dominant characters. The Duke and the Senators are specific to this
scene, and pervade the dialogue up to its end. In the second
segment, the JS divergence is maximal at line $230$, in the middle
of the first scene of Act II. This maximum can be ascribed to a
rather abrupt growth in the appearance frequency of Cassio. During
Act I and the first part of Act II, he plays an essentially marginal
role, with some $20$ lines. By the end of Act II, however, he has
had more than $70$ lines. In the third segment, the JS divergency
attains a sharp maximum at line $615$. This is again due to Cassio,
who has remained silent for most of Act III, but reappears with many
lines by the end of that act and the beginning of the following. At
the same time, Desdemona and Emilia, who had been on stage during
large sections of Act III, become considerably less frequent until
their reappearance by the middle of the Act IV. Finally, the maximum
at the fourth segment, line $1050$, coincides exactly with the last
utterance by Desdemona, at the fatal moment where she dies by his
jealous husband's hand.

Figures \ref{oth1} to \ref{oth3} show, as horizontal lines, the
average values and dispersions of the JS divergence for random
sequences with the same lengths and symbol frequencies as the
corresponding segments, calculated as explained in Section
\ref{anal}. All the maxima in the JS divergence are well above the
intervals $(\bar D - \sigma_D,\bar D + \sigma_D)$, indicating that
--at the analyzed levels-- the segmentation is detecting highly
significant non-random compositional differences.

Let us point out a few relevant facts. Note first that Cassio and
Desdemona are involved in the determination of many segmentation
points --at least, up to the segmentation level considered here.
This is due to the fact that, among the main characters, their
appearance frequencies varies most irregularly. Both are absent
during rather long sections, but participate actively when they are
on scene. On the other hand, the  ubiquitous, evil-doer Iago --the
second most frequent protagonist-- is evenly present over the whole
play. His role in defining contextual segments is therefore not
important at the considered levels. Secondly, note that, at the last
segmentation level (Figure \ref{oth3}), most transitions between
contiguous scenes and acts correspond either to a segmentation point
or to a local maximum in the JS divergence. Although local maxima
are not involved in the segmentation process, they also reveal
points with high compositional differences at both sides. At this
level, thus, the JS divergence is detecting the domains chosen by
Shakespeare  to divide his play. Finally, also in Figure \ref{oth3},
note the occurrence of many U-shaped structures, some of them
indicated by arrows. They correspond to dialogues between two
characters, whose alternate utterances give rise to this
characteristic profile. The insert in the figure is a close-up of
the main plot in one of these regions, showing the correspondingly
alternating values of the JS divergence. Such small-scale structures
are already clearly discerned at this segmentation level.

\section{Segmentation of a musical sequence}
\label{mus}

The relative frequency in the use of the different tones of the
twelve-tone musical scale is the primordial element determining what
in modern (post-Renaissance) western music is called the tonality of
a composition, or of one of its parts. Conditions of consonance and
dissonance of simultaneously sounding tones, as well as the
aesthetic viability of certain tone sequences, determine how often
they occur with respect to each other. In the tonality of C major,
for instance, the tone G (the so-called dominant of C) is expected
to occur much more frequently than G sharp (G\#), while the opposite
is true in E major (where G\# is part of the fundamental chord,
E$-$G\#$-$B) \cite{harm}. In connection with the present discussion,
tonality defines a contextual frame associated with the more or less
frequent occurrence of different tones. Tones are the perceptual
elements which define the tonal context \cite{mus}.

Since the early Baroque period, a systematic way for adding interest
to a musical composition has been to modify the tonality as the
composition progresses --a procedure called modulation. As a result
of modulation, successive parts of the composition correspond to
different tonal contexts. This procedure proved to be so
aesthetically convincing that certain musical forms (such as the
sonata; see below) developed standardized modulation motifs.

The segmentation of a symbolic sequence representing the succession
of tones in a musical composition should disclose the structure
associated with the transitions between tonal contexts. To show
this, we first identify each of the twelve tones as a different
symbol. Second, we must overcome the drawback that a composition is
not just a sequence of tones since, at a given time, many of them
may sound simultaneously. This can be solved by exploiting the
discreteness of time patterns in most musical styles, defined by
rhythm, to introduce a kind of coarse-grained discretization of the
succession of tones. Rhythm beats are grouped into bars (also called
measures), all of them usually containing the same number of beats.
Bars are the basic units underlying the rhythmic pattern. Every tone
can be unambiguously assigned to the bar where it begins. In this
way, it is possible to build a sequence with the number of
occurrences of each symbol in each bar in the succession. When
calculating the JS divergence, the relative frequencies of symbols
are obtained as the normalized sum of those numbers to the right and
to the left of the segmentation point.

\begin{figure}
\resizebox{\columnwidth}{!}{\includegraphics*{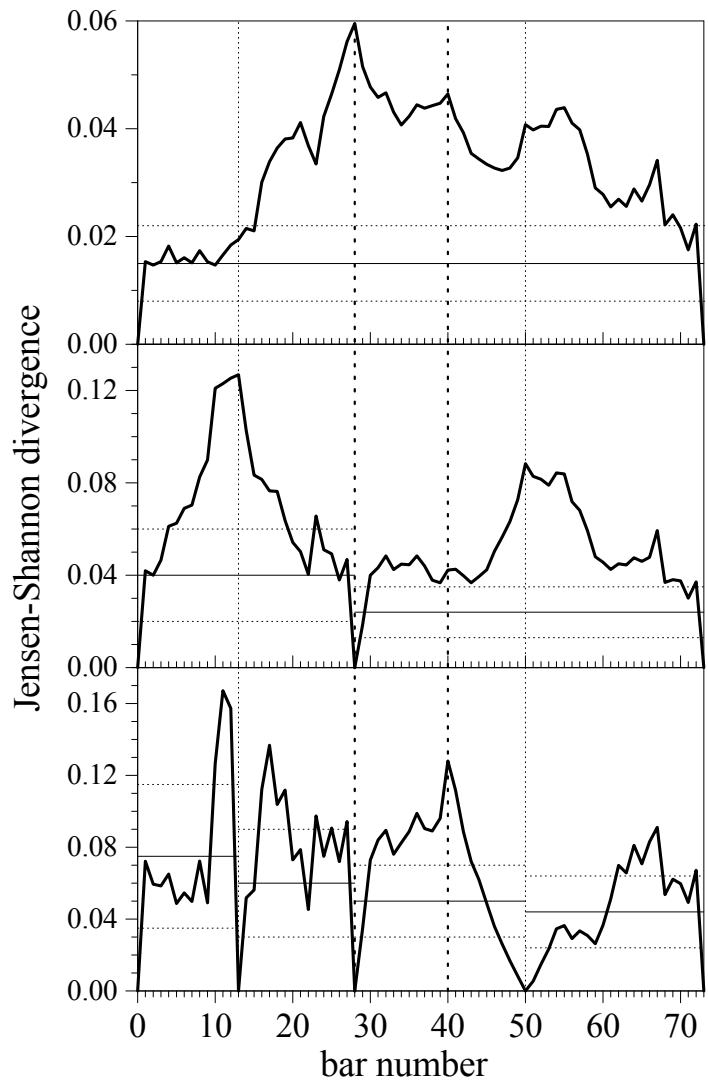}}
\caption{Jensen-Shannon divergence at the first three segmentation
levels, for the symbolic sequence obtained from Mozart's sonata in C
Major (K. 545, first movement). Vertical dotted lines correspond to
the division into the three sonata sections $ABA'$ (bold) and into
melodic themes (light). Horizontal lines are as in Figure
\ref{oth1}.} \label{k545}
\end{figure}

We apply the segmentation algorithm to a sequence obtained from the
first movement of the keyboard sonata in C major K.~545 by Mozart,
which spans $73$ bars. The sequence is built automatically from a
digital version of the composition in MIDI format \cite{mus,k545}.
The result of the three first segmentation steps is shown in Figure
\ref{k545}. As explained in the following, the analysis of the
musical score reveals that {\sl all} the segments detected by the
algorithm can be identified with well-defined domains of tonal
context in the composition.

The first movement of the sonata K.~545 has a standard ternary
structure --which, rather confusingly, is called sonata form--
usually denoted as $ABA'$. Section $A$ corresponds to the
exposition, where the main musical material of the composition is
introduced; $B$ is the development section, where the material is
elaborated in various forms; finally, in the recapitulation $A'$ the
material is again presented in its original form, usually with some
variations, leading to the conclusion. As shown in the upper panel
of Figure \ref{k545}, the JS divergence calculated over the whole
sequence of $73$ bars has a well-defined maximum at bar $28$, which
exactly coincides with the end of section $A$. In this composition,
a sudden tonality change, from G major to G minor, occurs at the
transition from section $A$ to $B$. This is detected by the JS
divergence as the most important boundary of frequency difference in
the use of tones.

Within the exposition section $A$, the standard sonata form
prescribes the presentation of two different melodic themes, in two
different tonalities: the main tonality of the composition (in our
case, C major) and, typically, its dominant (G major). The remaining
of the composition, sections $B$ and $A'$, can have a freer form. In
the first movement of K.~545, section $B$ is an elaboration of part
of the second theme presented in $A$ combined with new melodic
material, and progresses from the tonalities of G minor to D minor.
The recapitulation $A'$ re-introduces the first and the second theme
of section $A$ --but now in F major and C major, respectively--
before proceeding to the coda and ending the movement.

In the middle panel of Figure \ref{k545}, we see that the maximum of
the JS divergence in the first segment occurs at bar $13$. This is
the first bar of the second theme's exposition in section $A$. The
maximum thus coincides with the tonal transition between the two
themes, from C major to G major. In the second segment, the maximum
is at bar $50$, the last bar in the F major part of section $A'$.
Here, again, the JS divergence is detecting a tonality change, from
F major to C major.

At the next segmentation level, shown in the lower panel of Figure
\ref{k545}, the JS divergence reveals more subtle tonal structures.
Still, one of the maxima corresponds to a tonality change. In the
third segment, the maximum at bar $40$ coincides with the boundary
between sections $B$ and $A'$, with a modulation from D minor to F
major. At this level, thus, the three sonata sections become
separated from each other. The maximum in the first segment, at bar
$11$, discriminates between the bulk of the first theme, in C major,
and two bars where the modulation to the second theme's G major
takes place. In the second segment, the maximum is at bar $17$. This
divides the presentation of the simple second theme's melodic line
from the ensuing episode, a rather long series of arpeggio figures.
The tonal progression in this episode has the form of a sequence of
fifths, a well-known procedure by which the composition passes
between several, slightly different tonality contexts. This relative
tonal richness contrasts with the second theme's line. Finally, the
maximum in the fourth segment at bar $67$ separates the
re-exposition of the second time in section $A'$, in the tonality of
C major, from the coda. This last segment begins at bar $68$ with
the use of a rather special tone combination, a diminished seventh
cord, which --before the relaxation of the final bars-- establishes
a moment of tonal tension, clearly detected by JS divergence.

At the three levels, the maxima in the JS divergence are above the
intervals  $(\bar D - \sigma_D,\bar D + \sigma_D)$, shown in Figure
\ref{k545} by horizontal lines. Those intervals, however, are
relatively wider than in the case of the symbolic sequence
considered in Section \ref{lit}. This can be ascribed to the fact
that the whole musical sequence, as well as any of its segments, are
considerable shorter than those of literary origin, which enhances
the effect of fluctuations. Nevertheless, all the segmentations
performed on the musical sequence can be confidently associated with
significant divergence in the symbol frequencies.

\section{Conclusion}

We have illustrated the application of a segmentation algorithm,
based on the comparison of the compositional difference between
segments through the Jensen-Shannon divergence, to symbolic
sequences of literary and musical origin. The main aim has been to
test whether a method already used to detect structural
heterogeneities in arrays of symbols conveying information
--specifically, DNA sequences-- was able to produce any significant
result when applied to sequences derived from information processing
at a much higher level, such as in human languages. The answer is
positive: most of the segments resulting from the application of the
algorithm can be clearly identified with contextual domains inherent
to the standard structure of the literary and musical works from
where the analyzed sequences were extracted. Other segments, whose
interpretation requires effectively inspecting the changes in the
frequency of symbols along the sequence, point out more subtle
patterns, specific to the works in question. In all cases, however,
the segmentation procedure has amply fulfilled the significance
criterion used to discern long-range compositional divergence from
random-like local fluctuations.

The present segmentation algorithm has been able to detect, by means
of an unsupervised procedure, most of the structural division of a
theatrical play into acts and scenes, as designed by the play's
author. For the sequence of musical origin, it has been able to
disclose the same tonal patterns that would arise from the analysis
performed by a purposefully educated human. This suggests that such
analysis could be automatized, and eventually used as a helpful
instrument in such tasks as information processing, classification,
and retrieval in digitalized corpora and databases. It is however
clear that human intervention is expected at least at two stages.
First, an intelligent choice is needed for the translation of a
generic message into a symbolic sequence, which may depend on the
kind of contextual information to be extracted. Second, as
illustrated in the case of the sequence of literary origin, the full
interpretation of results may require their further elaboration and
comparison with information from other sources.

On the other hand, our contribution shows that significant results
can be obtained from an information-theoretical analysis of such
objects as literary or musical works which, traditionally, are
treated by means of less quantitative methods. The complementarity
of both approaches should emphasize the interdisciplinary interest
of the present study.

\section*{Acknowledgments}

I would like to thank Ana Majtey, from Universidad Nacional de
C\'or\-do\-ba, Argentina, for pointing out the Jensen-Shannon
divergence as a tool for the analysis of symbolic sequences.

\end{document}